\documentclass[runningheads]{llncs}
\usepackage{graphicx}
\usepackage{caption}
\usepackage[numbers,sort]{natbib}
\usepackage{tikz}
\usepackage{comment}
\usepackage{amsmath,amssymb} 
\usepackage{color}
\usepackage{float}

\usepackage{graphicx}
\usepackage{booktabs}
\usepackage{tikz}
\usepackage{multirow}

\usepackage[accsupp]{axessibility}  

\usepackage[width=122mm,left=12mm,paperwidth=146mm,height=193mm,top=12mm,paperheight=217mm]{geometry}

\newcommand{\fakepara}[1]{\vspace{2mm}\noindent\textbf{#1}\quad}

\begin{document}

\pagestyle{headings}
\mainmatter
\def\ECCVSubNumber{1415}  

\title{NeILF: Neural Incident Light Field for Physically-based Material Estimation} 

\titlerunning{NeILF}
%
\author{
Yao Yao \and
Jingyang Zhang \and
Jingbo Liu \and
Yihang Qu \and
Tian Fang \and
David McKinnon \and
Yanghai Tsin \and
Long Quan
}
\authorrunning{Yao et al.}

\institute{
Apple \hspace{15mm} HKUST
}
\maketitle

\begin{abstract}
We present a differentiable rendering framework for material and lighting estimation from multi-view images and a reconstructed geometry. In the framework, we represent scene lightings as the Neural Incident Light Field (NeILF) and material properties as the surface BRDF modelled by multi-layer perceptrons. Compared with recent approaches that approximate scene lightings as the 2D environment map, NeILF is a fully 5D light field that is capable of modelling illuminations of any static scenes. In addition, occlusions and indirect lights can be handled naturally by the NeILF representation without requiring multiple bounces of ray tracing, making it possible to estimate material properties even for scenes with complex lightings and geometries. We also propose a smoothness regularization and a Lambertian assumption to reduce the material-lighting ambiguity during the optimization. Our method strictly follows the physically-based rendering equation, and jointly optimizes material and lighting through the differentiable rendering process. We have intensively evaluated the proposed method on our in-house synthetic dataset, the DTU MVS dataset, and real-world BlendedMVS scenes. Our method is able to outperform previous methods by a significant margin in terms of novel view rendering quality, setting a new state-of-the-art for image-based material and lighting estimation.


\end{abstract}

\section{Introduction}

Material estimation from a set of sparse images is a challenging task in both computer vision and computer graphics. The problem is usually approached by inverse rendering, where the spatially-varying bidirectional reflectance distribution functions (SV-BRDFs) and lightings of the scene are jointly optimized by minimizing the rendering loss. 
However, the problem is hard to solve due to the complex form of the BRDF and the high-dimensional nature of scene illuminations. 
To mitigate the problem, previous methods usually apply simplified material and lighting models. For example, non-spatially varying BRDF~\cite{zhang2020physg} is applied for certain types of objects; approximated illuminations, such as co-located flash lights~\cite{nam2018practical,bi2020deep3d,bi2020deep,schmitt2020joint} and environment maps~\cite{zhang2020physg,zhang2020nerfactor,boss2021nerd,munkberg2021nvdiffrec,boss2021neuralpil}, are applied to reduce the complexity of the scene lighting. In most scenarios, special capturing devices or environments are required to assist the estimation, limiting these methods to real-world applications. As the result, a practical material estimator is still missing.

On the other hand, recent progress on neural representation has shown promising results for lighting modelling. NeRF~\cite{mildenhall2020nerf} jointly optimizes a neural radiance field and a density field, which has demonstrated great success for novel view synthesis. The surface light field is applied to model the outgoing light from the surface, which has been widely applied to neural surface reconstructions~\cite{yariv2020multiview,zhang2021learning}. Other methods further decompose observed lights into neural material properties and environmental lightings. However, similar to classical methods, they either use simplified lighting representations~\cite{bi2020deep,boss2021nerd,munkberg2021nvdiffrec,boss2021neuralpil}, or apply approximated occlusion and indirect light handling~\cite{zhang2020nerfactor,srinivasan2021nerv}. Until now, lighting modelling is still an open problem in image-based material estimation.

In this work, we address this long-standing problem by representing scene lightings as the neural incident light field.
Without losing generality, the proposed NeILF is capable of modelling lighting conditions of any static scenes. Also, occlusions and indirect lights could be naturally handled in the proposed framework without the need for tracing multiple bounces of rays. For material properties, we consider a simplified Disney BRDF model~\cite{burley2012physically} consisting of base color, roughness and metallic.
Implementation-wise, we use multi-layer perceptrons (MLPs) to represent both the incident light field and the BRDF. The NeILF network takes a 5D vector of location and incident direction as inputs, and returns as output a RGB value of the incident light; the material network takes a 3D location as input, and outputs a 5D vector of surface BRDF properties. Meanwhile, to reduce the ambiguity between the material and the scene lighting, we propose two regularization terms, namely the bilateral smoothness and the Lambertian assumption, to constrain the optimization of roughness and metallic.  Finally, we analyze similarities between NeILF for material estimation and NERF for novel view synthesis~\cite{mildenhall2020nerf}, providing readers an intuitive explanation of the difficulty and solvability of the problem.

We demonstrate in several datasets that our method significantly outperforms previous state-of-the-art in terms of novel view rendering accuracy. Our method is able to recover the surface BRDF even for scenes with complex lightings and geometries, which cannot be handled by previous environment map based methods. To summarize, main contributions of the paper include:
\vspace{-2mm}
\begin{itemize}
\item Representing scene lightings using the neural incident light field, where occlusions and indirect lights of the scene can be naturally handled.
\item A differentiable framework for joint material and lighting estimation, which significantly outperforms previous state-of-the-art in different datasets.
\item A bilateral smoothness and a Lambertian assumption to constrain the roughness and the metallic, reducing the material-lighting ambiguity during the network optimization.
\end{itemize}

\section{Related Works}

\subsection{The Rendering Equation} The rendering equation~\cite{kajiya1986rendering} computes the emitted radiance from a surface point $\mathbf{x}$ along a viewing direction $\boldsymbol{\omega_o}$:
\begin{equation}\label{eq:rendering}
	\begin{aligned}
		L_o(\boldsymbol{\omega_o}, \mathbf{x}) &= \int_{\Omega} f(\boldsymbol{\omega_o}, \boldsymbol{\omega_i}, \mathbf{x}) L_i(\boldsymbol\omega_i, \mathbf{x}) (\boldsymbol\omega_i\cdot \mathbf{n}) d\boldsymbol\omega_i,
	\end{aligned}
\end{equation}
where $\mathbf{n}$ is the normal of the surface, $ L_i $ is the incoming light from direction $\boldsymbol\omega_i$, and $f$ is the BRDF function to describe the the reflectance property, which is usually decomposed into a diffuse term and a specular term $f = f_{d} + f_{s}$. The integration is performed over all incident direction $\boldsymbol\omega_i $ on the hemisphere $ \Omega $ where $ \boldsymbol\omega_i\cdot \mathbf{n} >0 $.  

The goal of material estimation is to recover continues functions of the scene lighting $L_i$ and the BRDF property $f$ in the above equation.  Due to the complex form of the scene lighting and the material property, it is crucial to select suitable representations for $L_i$ and $f$. In this paper, we propose to use a neural incident light field to model $L_i$ (Sec.~\ref{sec:neilf}), and apply a simplified Disney BRDF~\cite{burley2012physically} model to approach the BRDF $f$ (Sec.~\ref{sec:brdf}). Below we give a brief review on the physically-based material estimation from multi-view images.

\subsection{Differentiable Rendering} 

Unlike classical approaches that recover 3D scene parameters in a forward reconstruction manner, differentiable rendering~\cite{kato2020differentiable,Azinovic_2019_CVPR} inverses the rendering process in graphics, and optimize all parameters by minimizing the difference between rendered and input images. 

Recently, the technique has been combined with neural representations and has shown promising results for image-based 3D problems. NeRF~\cite{mildenhall2020nerf} and follow-up works~\cite{liu2020neural,martin2021nerf,zhang2020nerf++} decouple a 3D scene into a density field and a radiance field. Other methods also apply implicit functions to model different geometry~\cite{liu2019learning,liu2020dist,zhang2021learning} and appearance representations~\cite{niemeyer2020differentiable,yariv2020multiview}. In another line of works, the received radiance is further decomposed into BRDF properties and input light sources~\cite{zhang2020physg,zhang2020nerfactor,boss2021nerd,srinivasan2021nerv}. Our method follows this practice and applies a neural BRDF model to approach the material property of the surface.

\subsection{Material and Lighting Estimation} 

Due to the difficulty of joint material and lighting estimation, previous methods usually apply additional sensors or controlled lightings to facilitate the optimization process. For example, additional sensors~\cite{guo2019relightables,azinovic2019inverse,park2020seeing}, co-located flash lights~\cite{nam2018practical,bi2020deep3d,bi2020deep,schmitt2020joint}, or turn-table settings~\cite{dong2014appearance,xia2016recovering} are applied to capture image or scene lightings. Moreover, simplified material and lighting models, e.g., the non-spatially varying BRDF~\cite{zhang2020physg} or approximated illuminations of environment maps~\cite{zhang2020physg,zhang2020nerfactor,boss2021nerd}, are applied to reduce the complexity of the problem. NeRV~\cite{srinivasan2021nerv} 
introduces the visibility field to model indirect lights, however, requires environmental lightings to be known in advance. Nevertheless, such mitigations will inevitably limit these methods to real-world applications. In contrast, our method applies a unified incident light field to represent different light sources in the scene, and is capable of jointly estimating material and lighting under any lighting conditions.

\begin{figure}[t]
	\centering
	\includegraphics[width=1.0\linewidth]{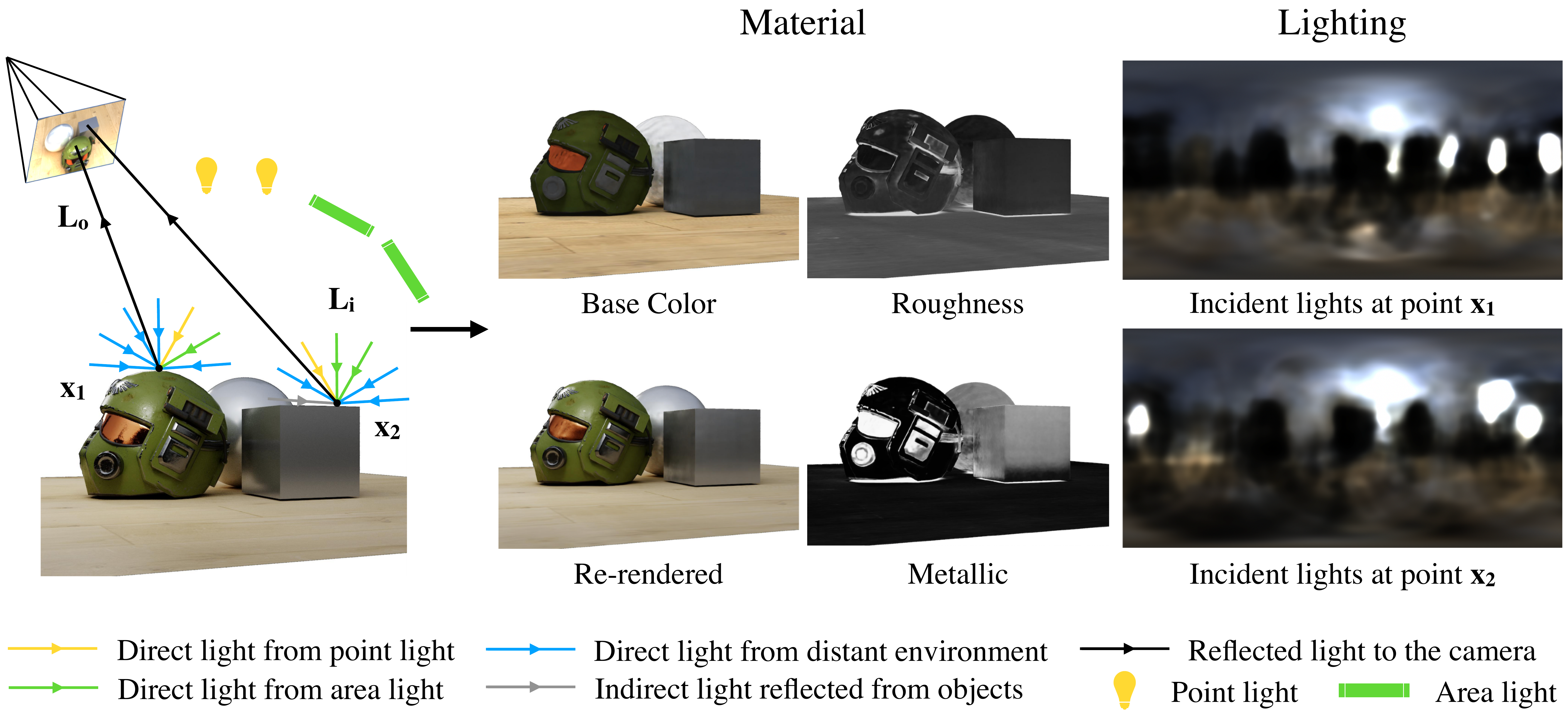}
	\caption{Illustrations of the proposed method and our material and lighting estimation results. NeILF is capable of modelling the joint illumination of direct/indirect lights from different sources. Estimated incident lights at point $\mathbf{x_1}$ and point $\mathbf{x_2}$ well explain the mixed lighting of the scene, including an environment map with high-radiance sun light, two near-range point lights, and two near-range area lights.}
	\label{fig:pipeline}
\end{figure}

\section{Method}

\subsection{Neural Incident Light Field}\label{sec:neilf}

One of the keys to invert the rendering equation is to model the incoming light $L_i$ in a correct way. Ideally, $L_i$ should take into account 1) \textit{direct lights} from light sources in the scene, 2) \textit{occlusions} that block the surface point from receiving direct lights, and 3) \textit{indirect lights} that are reflected from other surface points. However, each of the three components is hard to model. Previous methods~\cite{li2018learning,zhang2020physg,zhang2020nerfactor,boss2021nerd} usually approximate direct lights as an environment map and hardly handle indirect lights as they require multi-bounce raytracing.


In contrast, we formulate incoming lights in the scene directly as a neural incident light field, where an MLP takes a point location $\mathbf{x}$ and an incident direction $\boldsymbol\omega$ as inputs, and returns an incident light radiance $L$ as output:
\begin{equation}
	\begin{aligned}
		\mathbb{L}:  \{\mathbf{x}, \boldsymbol\omega\} \to \mathbf{L}.
	\end{aligned}
\end{equation}

Without losing generality, the proposed NeILF representation is capable of modelling the joint illumination effect of direct/indirect lights and occlusions of \textit{any static scenes}. An illustration is shown in Fig.~\ref{fig:pipeline}. Compared with the commonly used environment map, NeILF is able to handle the spatially-varying illumination effect, making it possible to estimate material for scenes with complex geometries and lightings. 

\subsection{Simplified Disney BRDF}\label{sec:brdf}
In this section, we describe the BRDF representation used in the proposed framework. We apply a simplified Disney principled BRDF model, where the BRDF of a surface point $\mathbf{x}$ is parameterized by a base color $ \mathbf{b}(\mathbf{x}) \in [0,1]^3 $, a roughness $ r(\mathbf{x}) \in [0,1] $ and a metallic $ m(\mathbf{x}) \in [0,1] $, which is a subset of the full Disney model~\cite{burley2012physically}. Similar to the neural incident lighting field, BRDF parameters are also stored using multi-layer perceptrons:
\begin{equation}
	\begin{aligned}
		\mathbb{B}: \mathbf{x} \to \{\mathbf{b}, r, m\},
	\end{aligned}
\end{equation}
where the MLP takes a 3D surface point $\mathbf{x}$ as input, and returns the 5-channel BRDF parameters as output. Note that other representations, e.g. UV atlas or per-vertex BRDF parameters, can also be applied. Here we choose the neural representation because it has been proven to be effective for modelling continuous functions in 3D space~\cite{mildenhall2020nerf,zhang2020physg}, and its derivative can be easily and analytically derived for our regularization computation (Sec.~\ref{sec:regularization}). 

\fakepara{The rendering equation} Given the BRDF parameterization, we now describe the concrete formulation of $f$ in Equation~\ref{eq:rendering}. In the following equations, we omit notations of surface point $\mathbf{x}$ and normal $\mathbf{n}$ as the geometry of the scene is assumed to be given. The diffuse term can be calculated as $f_d = \frac{1 - m}{\pi} \cdot \mathbf{b}$, and the specular term as:
\begin{equation}\label{eq:f_s}
	\begin{aligned}
		f_s(\boldsymbol\omega_o, \boldsymbol\omega_i) &= \frac{D(\mathbf{h}; r) \cdot F(\boldsymbol\omega_o, \mathbf{h}; \mathbf{b}, m) \cdot G(\boldsymbol{\omega}_i, \boldsymbol\omega_o, \mathbf{h}; r)}{(\mathbf{n} \cdot \boldsymbol\omega_i) \cdot (\mathbf{n} \cdot \boldsymbol\omega_o)}, 
	\end{aligned}
\end{equation}
where $\mathbf{h}$ is the half vector between the incident direction $\boldsymbol\omega_i$ and the viewing direction $\boldsymbol\omega_o$. The first term $D$ is the normal distribution function of the microfacets in the surface. It is related to the roughness $r$ and we use Spherical Gaussians to model this function as in previous methods~\cite{wang2009all,zhang2020physg}. The second Fresnel term $F$ models the portion of light that can be reflected from the surface, which is determined by the surface metallic $m$ and the base color $\mathbf{b}$. The final geometry term $G$ handles the shadow and occlusion of the microfacets, which is parameterized on the roughness $r$ and is approximated using the GGX distribution~\cite{walter2006ggx}. Details of $D$, $F$ and $G$ are provided in the supplementary material.

\subsection{Material-Lighting Ambiguity and Regularizations}\label{sec:regularization}

While the Disney BRDF and incident light field are capable of representing materials and lightings of different scenes, jointly optimizing both would inevitably lead to ambiguous solutions among them. One degenerate case could be that we can force a pure reflective BRDF to all surface points, and then only optimize the incident lights to adjust the input image. Theoretically, we can still find a perfect solution that fits the given BRDF and input images: for each 3D point, whenever there is a visible camera, we set its mirror symmetric incident light equal to the viewing out-going light, and set other incident lights equal to zero. It is also reported in previous work~\cite{pont2006material} that even human observers cannot distinguish the two confounded components from only image observations. 

In the proposed framework, we can still manage to recover reasonable material and lighting results as MLPs can implicitly enforce a spatial smoothness constraint~\cite{zhang2020nerf++} on the two components. However, for robust material and lighting estimation, additional regularizations are desired. In this paper, we propose two regularizations for roughness $r$ and metallic $m$:

\fakepara{Bilateral Smoothness} We encourage $r$ and $m$ not to change rapidly in space, and the gradient of the input image $\mathbf{I}$ can be used as a hint to guide the smoothing process. Thus, we define the bilateral smoothness cost of $r$ and $m$ as:
\begin{equation}
	l_{smooth} = \frac{1}{|S_I|} \sum_{\mathbf{p}\in S_I} (\| \nabla_{\mathbf{x}} r(\mathbf{x}_{\mathbf{p}}) \| + \| \nabla_{\mathbf{x}} m(\mathbf{x}_{\mathbf{p}}) \|) e^{-\| \nabla_{\mathbf{p}} \mathbf{I}(\mathbf{p}) \|},
\end{equation}
where $S_I$ is the set of all sampled pixels and $\mathbf{x_p}$ is the corresponding 3D point of the sampled pixel $\mathbf{p}$. The image gradient $\nabla_{\mathbf{p}} \mathbf{I}(\mathbf{p})$ can be pre-calculated from the input image, and the roughness gradient $\nabla_{\mathbf{x}} r(\mathbf{x}_{\mathbf{p}})$ and metallic gradient $\nabla_{\mathbf{x}} m(\mathbf{x}_{\mathbf{p}})$ can be derived analytically by back-propagating the neural network. 

\fakepara{Lambertian Assumption} We also assume that all surfaces tend to be Lambertian if no view-dependent lighting is observed, which leads to high roughness and low metallic, and we define the Lambertian cost as: 
\begin{equation}
	\begin{aligned}
		l_{lambertian} &= \frac{1}{|S_I|} \sum_{\mathbf{p}\in S_I} (|r(\mathbf{x_p})-1| + |m(\mathbf{p})| ).
	\end{aligned}
\end{equation}

The proposed two regularizations will be minimized during network training. It is noteworthy that the two losses may not necessarily improve quantitative results as they are heuristically defined for robust material and lighting estimation. We show in a later ablation study that the bilateral smoothness will lead to visually much more pleasing results for real-world reconstructions.

\subsection{Loss}

Similar to other differentiable rendering framework, we compute the L1 loss between the rendered image and the input image:
\begin{equation}\label{eq:l1}
	\begin{aligned}
        l_{image} = \frac{1}{|S_I|} \sum_{\mathbf{p}\in S_I} \| \mathbf{I}(\mathbf{p}) - L_o(\mathbf{x_p}, \boldsymbol\omega_o) \|_1.
	\end{aligned}
\end{equation}
The final loss of the proposed system is a weighted sum of the image loss and the two regularization losses: $l = l_{image} + w_s l_{smooth} + w_l l_{lambertian}$, 
where the two weights are empirically set to $w_s = 10^{-4}$ and $w_l = 10^{-3}$ in all our experiments.

\section{Implementations}

\subsection{Sphere Sampling}\label{sec:sample}

To compute $L_o$ using a finite number of incident lights, we need to discretize Equation~\ref{eq:rendering} as: $L_o(\boldsymbol{\omega_o}, \mathbf{x}) = \sum_{i \in S_L} f(\boldsymbol{\omega_o}, \boldsymbol{\omega_i}, \mathbf{x}) L_i(\boldsymbol\omega_i, \mathbf{x}) (\boldsymbol\omega_i\cdot \mathbf{n}) \cdot A(\boldsymbol\omega_i)$,
where $S_L$ is the set of incident lights sampled for point $\mathbf{x}$ and $A(\boldsymbol\omega_i)$ is the solid angle that corresponds to the incident light. In computer graphics, randomized Monto-Carlo Samplings are usually applied in ray-tracing, and the solid angle $A(\boldsymbol\omega_i)$ is approximated by the probability distribution $P(\boldsymbol\omega_i)$ of ray samples. 

However, in differentiable rendering, it is critical to accurately compute the solid angle $A(\boldsymbol\omega_i)$ for each light sample as we need to correctly pass loss gradients to network parameters. We found that using random sampling and approximating $A(\boldsymbol\omega_i)$ as the probability distribution $P(\boldsymbol\omega_i)$ will lead to erroneous BRDF results. Thus, we apply a fixed Fibonacci sampling over the half sphere to get all samples. In this case, $A(\boldsymbol\omega_i) = \frac{2 \pi}{|S_L|}$ and the rendering equation becomes:
\begin{equation}\label{eq:discretization}
	\begin{aligned}
		L_o(\boldsymbol{\omega_o}, \mathbf{x}) &= \frac{2 \pi}{|S_L|} \sum_{i \in S_L} f(\boldsymbol{\omega_o}, \boldsymbol{\omega_i}, \mathbf{x}) L_i(\boldsymbol\omega_i, \mathbf{x}) (\boldsymbol\omega_i\cdot \mathbf{n}).
	\end{aligned}
\end{equation}

\subsection{Learned HDR-LDR Mapping}\label{sec:ldr}

For real-world datasets with low dynamic range (LDR) images, we need to convert the high dynamic range (HDR) output from our renderer to LDR before computing the image loss. As such transformation is unavailable in previous MVS datasets, we apply a learned HDR-LDR mapping to mimic the conversion in our network. Note that linear transformations, including exposure and white balance, can be embedded into the incident light. Thus, we only explicitly model the gamma correction with a learnable parameter:
\begin{equation}\label{eq:ldr}
	\begin{aligned}
		L_o^{LDR} = (L_o^{HDR})^{\gamma}.
	\end{aligned}
\end{equation}

\subsection{Training Details}\label{sec:training}

We use an 8-layer Siren \cite{sitzmann2020implicit} with feature size of 512 and a skip connection in the middle to represent the BRDF MLP. Also, the positional encoding~\cite{martin2021nerf} is applied to further strengthen the network. The NeILF MLP shares the same implementation as BRDF, except that 1) the feature size is downsized to 128 to reduce the VRAM usage and 2) the last layer activation function is changed from \textit{tanh} to \textit{exp} in order to guarantee non-negative and unbounded light intensities.

In the experiment, we use $|S_L|=128$ incident lights to compute the output radiance during training, and use $|S_L|=256$ incident lights to evaluate the rendered image during testing. For each training iteration, we randomly sample 16000 pixels from all images, and the network is optimized for a total of 15000 iterations. The Adam optimizer~\cite{kingma2015adam} with an initial learning rate of $10^{-3}$ is applied in our network, and the learning rate is scaled down by $ \sqrt{0.1} $ at 5000 and 10000 iterations. The training process takes around 1.5 hours to finish on a Tesla V100 GPU and the VRAM consumption is around 30 GB.

\section{Experiments}

\subsection{Baseline Methods}\label{sec:baseline}
We compare our method with the following baselines:

\fakepara{PhySG$^*$} Firstly, we consider the recent PhySG \cite{zhang2020physg} for material estimation. The original PhySG jointly optimizes the non-spatially varying BRDF, the environment map, and the geometry of the object. To fairly compare with the method, we fix the given geometry and optimize only the uniform BRDF and the environment map. 

\fakepara{SG-Env} This baseline is another variant of PhySG~\cite{zhang2020physg}. Compared with PhySG$^*$, SG-Env applies a SV-BRDF model and a slightly different rendering formulation (the same $f_d$, $D$ and $G$ as ours). We use this baseline to directly compare NeILF with the SG environment map representation.

\fakepara{Pix-Env} Compared with SG-Env, this baseline uses a 2D image of resolution $32\times16$ to represent the environment map, which can be viewed as a variant of NeRFactor~\cite{zhang2020nerfactor} but without visibility handling. We sample all pixels in the environment image to compute the rendering equation, where pixel values are optimized directly during training. 

\fakepara{Ne-Env} Lastly, we compare our method with the neural environment map representation. This baseline shares the same implementation of the proposed NeILF, except that the positional input in the incident light field is omitted such that the incident light is only related to the incoming direction: $\{\boldsymbol\omega\} \to \mathbf{L}$. 

\begin{table}[]
\caption{Quantitative results on Synthetic scenes. We compare the proposed NeILF with four baseline methods described in Sec.~\ref{sec:baseline} using PSNR scores. Our method generates consistently the best novel view rendering for all scenes. Also, our method produces significantly better BRDF results than other methods if multiple objects and mixed lightings are given.}
\label{tab:syn}
\resizebox{\textwidth}{!}{%
\begin{tabular}{cc|ccc|cccccc}
\specialrule{.2em}{.1em}{.1em}
\multicolumn{2}{c|}{Scene Geometry}                        & \multicolumn{3}{c|}{Single-helmet}                  & \multicolumn{6}{c}{Combined-objects}                                                                                     \\ \hline
\multicolumn{2}{c|}{Scene Lighting}                        & Env-city & Env-studio     & Env-castel     & Env-city       & Env-studio     & \multicolumn{1}{c|}{Env-castel}     & Mix-city     & Mix-studio   & Mix-castel   \\ \hline
\multicolumn{1}{c|}{\multirow{5}{*}{Base Color}} & PhySG$^*$\cite{zhang2020physg}  & 13.43             & 14.87          & 13.95          & 15.01          & 16.96          & \multicolumn{1}{c|}{16.13}          & 14.16          & 12.43          & 14.29          \\
\multicolumn{1}{c|}{}                            & SG-ENV\cite{zhang2020physg}  & \textbf{20.61}    & \textbf{18.45} & 15.99          & \textbf{22.38} & \textbf{20.74} & \multicolumn{1}{c|}{\textbf{22.21}} & 16.92          & 13.16          & 16.69          \\
\multicolumn{1}{c|}{}                            & Pix-ENV\cite{zhang2020nerfactor} & {14.20}    & 13.37          & 12.18          & 13.18          & 15.09          & \multicolumn{1}{c|}{7.94}           & 12.16          & 11.44          & 11.79          \\
\multicolumn{1}{c|}{}                            & Ne-ENV  & 13.43             & 12.65          & 12.45          & 11.68          & 11.98          & \multicolumn{1}{c|}{7.66}           & 11.87          & 10.90          & 9.54           \\
\multicolumn{1}{c|}{}                            & Ours    & 16.36             & 16.36          & \textbf{18.28} & 15.59          & 15.48          & \multicolumn{1}{c|}{12.95}          & \textbf{17.39} & \textbf{16.88} & \textbf{17.37} \\ \hline
\multicolumn{1}{c|}{\multirow{5}{*}{Metallic}}   & PhySG$^*$\cite{zhang2020physg}  & 7.57              & 7.48           & 7.83           & 8.72           & 7.97           & \multicolumn{1}{c|}{8.35}           & 8.67           & 8.95           & 8.76           \\
\multicolumn{1}{c|}{}                            & SG-ENV\cite{zhang2020physg}  & \textbf{21.19}    & \textbf{21.31} & \textbf{21.79} & 17.01          & 16.40          & \multicolumn{1}{c|}{\textbf{16.39}} & 15.44          & 14.25          & 14.49          \\
\multicolumn{1}{c|}{}                            & Pix-ENV\cite{zhang2020nerfactor} & 14.28             & 18.03          & 18.44          & 16.15          & 15.61          & \multicolumn{1}{c|}{11.42}          & 15.27          & 14.24          & 14.84          \\
\multicolumn{1}{c|}{}                            & Ne-ENV  & 9.31              & 17.40          & 6.15           & 15.86          & 16.10          & \multicolumn{1}{c|}{11.42}          & 15.43          & 15.35          & 15.49          \\
\multicolumn{1}{c|}{}                            & Ours    & 17.79             & 18.52          & 16.82          & \textbf{18.22} & \textbf{19.11} & \multicolumn{1}{c|}{10.29}          & \textbf{18.42} & \textbf{18.43} & \textbf{17.34} \\ \hline
\multicolumn{1}{c|}{\multirow{5}{*}{Roughness}}  & PhySG$^*$\cite{zhang2020physg}  & 6.91              & 11.88          & 6.75           & 6.62           & 11.29          & \multicolumn{1}{c|}{6.22}           & 6.27           & 6.83           & 6.14           \\
\multicolumn{1}{c|}{}                            & SG-ENV\cite{zhang2020physg}  & 14.77             & 15.87          & 9.64           & 9.61           & 17.64          & \multicolumn{1}{c|}{9.74}           & 8.77           & 12.58          & 9.14           \\
\multicolumn{1}{c|}{}                            & Pix-ENV\cite{zhang2020nerfactor} & 13.88             & 14.36          & 15.92          & 15.55          & 13.95          & \multicolumn{1}{c|}{13.06}          & 16.11          & 14.12          & 13.41          \\
\multicolumn{1}{c|}{}                            & Ne-ENV  & 11.95             & 14.84          & 9.26           & 15.56          & 14.48          & \multicolumn{1}{c|}{12.94}          & 16.20          & 14.43          & 14.14          \\
\multicolumn{1}{c|}{}                            & Ours    & \textbf{16.13}    & \textbf{16.19} & \textbf{17.16} & \textbf{17.48} & \textbf{18.30} & \multicolumn{1}{c|}{\textbf{13.40}} & \textbf{17.05} & \textbf{16.27} & \textbf{16.44} \\ \hline
\multicolumn{1}{c|}{\multirow{5}{*}{Rendering}}  & PhySG$^*$\cite{zhang2020physg}  & 24.59             & 24.77          & 26.52          & 24.82          & 25.65          & \multicolumn{1}{c|}{27.24}          & 24.38          & 24.04          & 25.81          \\
\multicolumn{1}{c|}{}                            & SG-ENV\cite{zhang2020physg}  & 29.73             & 29.86          & 32.13          & 31.01          & 29.46          & \multicolumn{1}{c|}{32.34}          & 27.20          & 25.88          & 27.70          \\
\multicolumn{1}{c|}{}                            & Pix-ENV\cite{zhang2020nerfactor} & 29.28             & 29.03          & 31.09          & 30.81          & 29.06          & \multicolumn{1}{c|}{32.30}          & 27.88          & 25.85          & 27.97          \\
\multicolumn{1}{c|}{}                            & Ne-ENV  & 28.60             & 29.56          & 29.76          & 30.75          & 29.07          & \multicolumn{1}{c|}{32.05}          & 28.07          & 26.01          & 28.33          \\
\multicolumn{1}{c|}{}                            & Ours    & \textbf{31.57}    & \textbf{30.84} & \textbf{34.43} & \textbf{33.77} & \textbf{31.07} & \multicolumn{1}{c|}{\textbf{35.28}} & \textbf{31.11} & \textbf{28.59} & \textbf{32.11} \\
\specialrule{.2em}{.1em}{.1em}
\end{tabular}%
}
\end{table}

\begin{figure}
	\centering
	\includegraphics[width=1.0\linewidth]{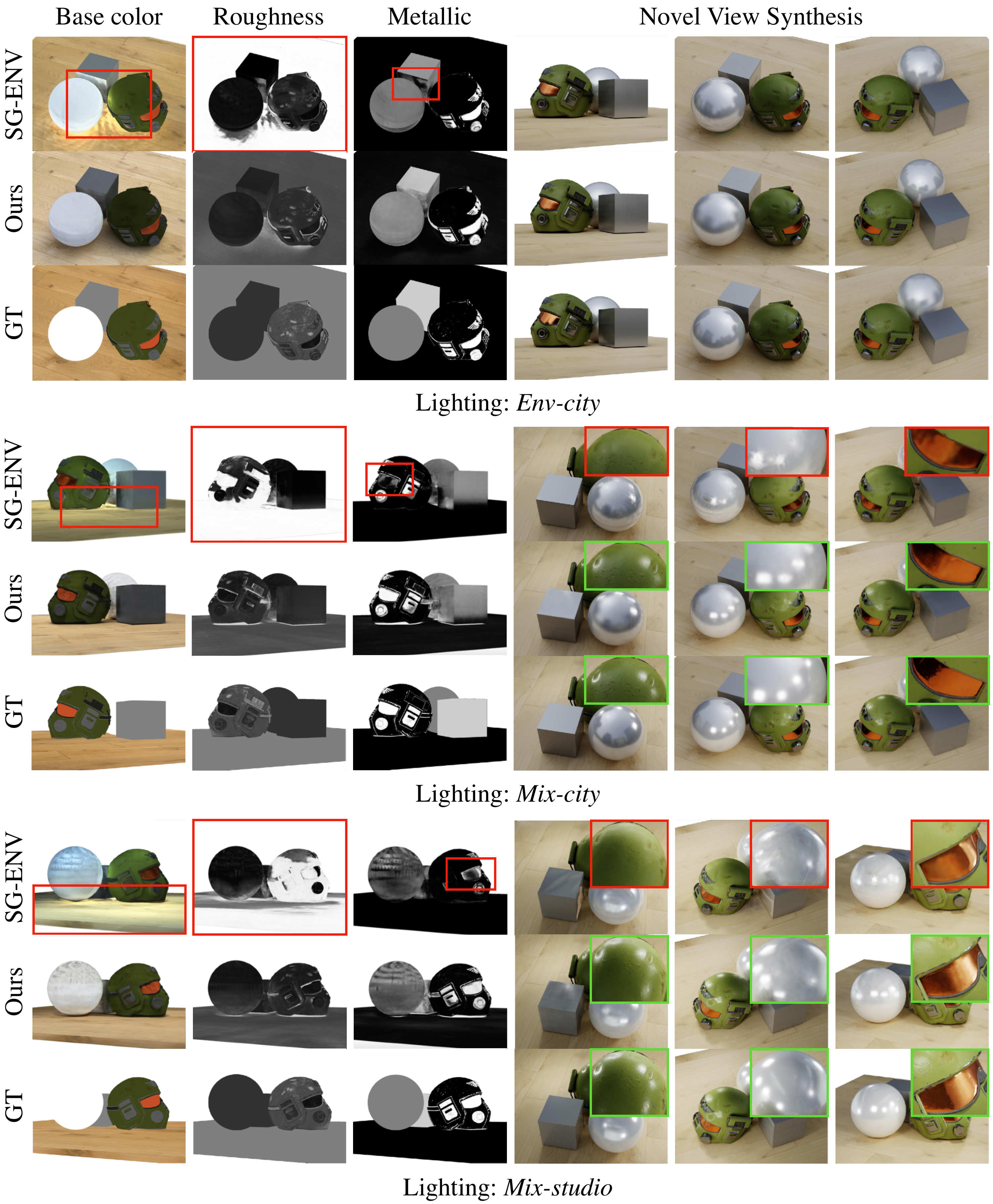}
	\caption{Comparative results on BRDF estimation and novel view synthesis on the synthetic dataset. From left to right are images of \textit{base color}, \textit{roughness}, \textit{metallic} and synthesized testing views. Our method is able to generate high-quality BRDF and novel view synthesis results under different lighting conditions. In contrast, the environment map based SG-ENV~\cite{zhang2020physg} produces noisy BRDF outputs especially in occluded regions. And also, high lights are wrongly recovered in novel view renderings if mixed lightings occur.}
	\label{fig:syn}
\end{figure}

\subsection{Benchmark on Synthetic Scenes}

To quantitatively evaluate our method under different lighting conditions, we generate a set of synthetic data and compare our method with the above baselines.

\fakepara{Data Preparation} The synthetic dataset contains three objects and their combinations: a single rough metallic sphere, a single rough metallic cube, and a helmet with spatially variant materials. The objects are placed on a plane to model the real-world object capture. We also create six lighting conditions to lit the objects, including three environment maps and three mixed lightings: 
\begin{itemize}
    \item \textit{Env–city}: an environment map of a city;
    \item \textit{Env–studio}: an environment map of a studio;
    \item \textit{Env–castel}: an environment map of a castel;
    \item \textit{Mix–city}: Env–city plus two point lights and two area lights;
    \item \textit{Mix–studio}: Env–studio plus two point lights and two area lights;
    \item \textit{Mix–castel}: Env–castel plus two point lights and two area lights.
\end{itemize}
Each scene contains 96 images, where 87 images are used for training and the left out 9 images are used for evaluation. The image trajectory forms three loops around the object at altitude 0, 22.5 and 45 and the image resolution is set to $1600 \times 1200$. We use Blender~\cite{blender} to render the HDR images by ray tracing. Position maps and normal maps at all viewpoints are rendered to serve as the geometry input for the system. Meanwhile, per-view ground truth base color, metallic and roughness maps are provided for quantitative evaluation. 

\fakepara{Results} We use the PSNR score as our evaluation metric. Quantitative comparisons on 1) base color, 2) metallic, 3) roughness and 4) novel view synthesis are shown in Table~\ref{tab:syn}. Our method consistently outperforms other methods with a large margin in terms of the novel view rendering quality. For material estimation, we found that if single objects and environment map light sources are given, SG-Env~\cite{zhang2020physg} is able to generate comparable results with ours. However, if multiple objects or mixed light sources are given, its quality will drop significantly. This is because the environment map representation cannot model mixed light sources of point and area lights. Also, indirect lights and occlusions within multiple objects are not handled by SG-Env. In contrast, our NeILF representation can robustly deal with mixed lightings and complex scene geometries. Qualitative results are shown in Fig.~\ref{fig:syn}.

\begin{table}[t!]
\caption{Quantitative results on DTU~\cite{jensen2014large} and BlendedMVS~\cite{yao2020blendedmvs} Datasets. The table shows PSNR scores of novel view renderings of test images. Our method consistently outperforms the other methods in terms of the rendering quality.}
\label{tab:real}
\resizebox{\textwidth}{!}{%
\begin{tabular}{c|ccccc|cccccc}
\specialrule{.2em}{.1em}{.1em}
             & \multicolumn{5}{c|}{DTU~\cite{jensen2014large}}                                                           & \multicolumn{6}{c}{BlendedMVS~\cite{yao2020blendedmvs}}                                                                      \\ \hline
             & scan-1         & scan-11        & scan-37        & scan-75        & scan-97        & bull           & cam            & dog            & gold           & statue         & stone          \\ \hline
PhySG$^*$~\cite{zhang2020physg}       & 20.40          & 20.78          & 20.30          & 16.03          & 19.86          & 21.64          & 18.11          & 20.70          & 19.06          & 19.74          & 21.22          \\
SG-ENV~\cite{zhang2020physg}       & 22.18          & 21.56          & {21.71} & 18.06          & {21.09} & {22.51} & {20.14} & 22.06          & 19.44          & 20.79          & 22.31          \\
Pix-ENV~\cite{zhang2020nerfactor}       & 23.61          & 23.61          & 22.68          & 19.54          & 21.52          & 21.58          & 19.98          & 21.36          & 19.28          & 20.46          & 22.89          \\
Ne-ENV       & 23.77          & 23.79          & 22.87          & 19.52          & 21.51          & 22.17          & 20.17          & 21.73          & 19.66          & 20.55          & 23.08          \\
NeILF (Ours) & \textbf{24.79} & \textbf{24.33} & \textbf{24.44} & \textbf{23.46} & \textbf{23.96} & \textbf{24.93} & \textbf{22.10} & \textbf{22.36} & \textbf{20.80} & \textbf{21.51} & \textbf{24.22} \\ 
\specialrule{.2em}{.1em}{.1em}
\end{tabular}%
}
\end{table}
\begin{figure}[t!]
	\centering
	\includegraphics[width=1.0\linewidth]{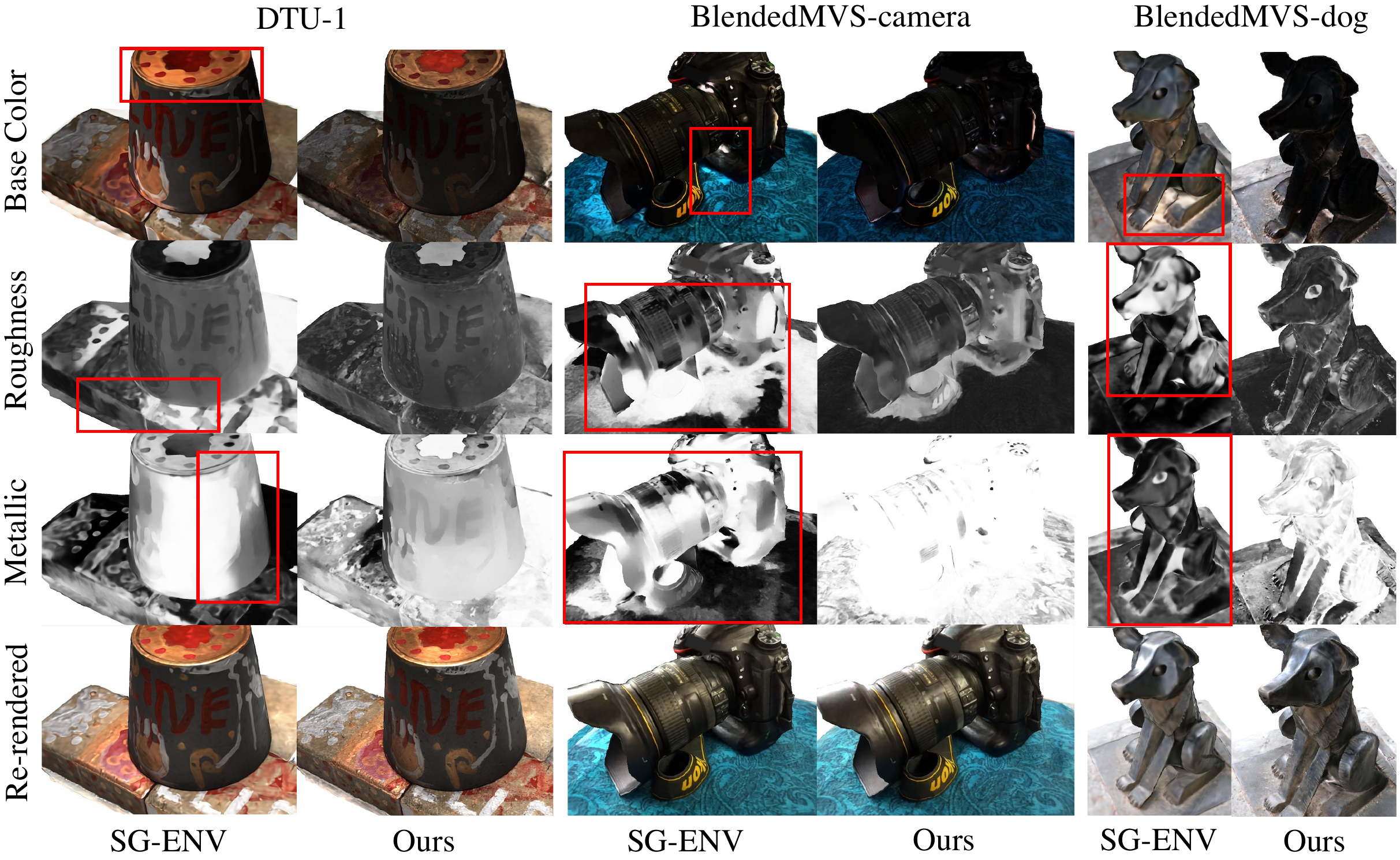}
	\caption{Qualitative results on DTU~\cite{jensen2014large} and BlendedMVS~\cite{yao2020blendedmvs} datasets. Our method successfully removes high lights in the base color and produces visually plausible results of roughness and metallic. Note that brand names in BlendedMVS-camera have been masked out from images.}
	\label{fig:real}
\end{figure}

\subsection{Test on Real-world Scenes}
We then test our method on two real-world datasets, namely DTU~\cite{jensen2014large} and BlendedMVS~\cite{yao2020blendedmvs} datasets. DTU dataset is captured in a lab setting with a fixed lighting and camera trajectory, while BlendedMVS contains a variety of indoor and outdoor scenes captured by different users. As the two datasets provide only LDR images, the learned HDR-LDR mapping described in Sec.~\ref{sec:ldr} is applied for the loss computation. For each scene, we select 5 images for testing and use remaining images for network training. 

It is noteworthy that unlike in the synthetic dataset, here we use multi-view stereo methods to generate the geometry input rather than directly using the ground truth. For DTU datatset, we use Vis-MVSNet~\cite{zhang2020visibility} to generate the dense 3D point cloud and SPSR~\cite{kazhdan2013screened} to recover the mesh surface. For BlendedMVS dataset, we use original images and the provided reference mesh geometry as our inputs. By doing so, our method can be viewed as an extension to nowadays 3D reconstruction pipelines.

Quantitative results are shown in Table~\ref{tab:real} and qualitative results compared with SG-ENV are shown in Fig.~\ref{fig:real}. The proposed method produces both the best rendering PSNR and the most visually pleasing BRDF in all selected scenes. We believe the proposed method can be integrated into traditional 3D reconstruction pipelines~\cite{furukawa2009accurate,schonberger2016pixelwise,yao2018mvsnet,kazhdan2006poisson,hiep2009towards} for relightable mesh model reconstruction. 

\begin{figure}[t]
    \centering
    \includegraphics[width=1\linewidth]{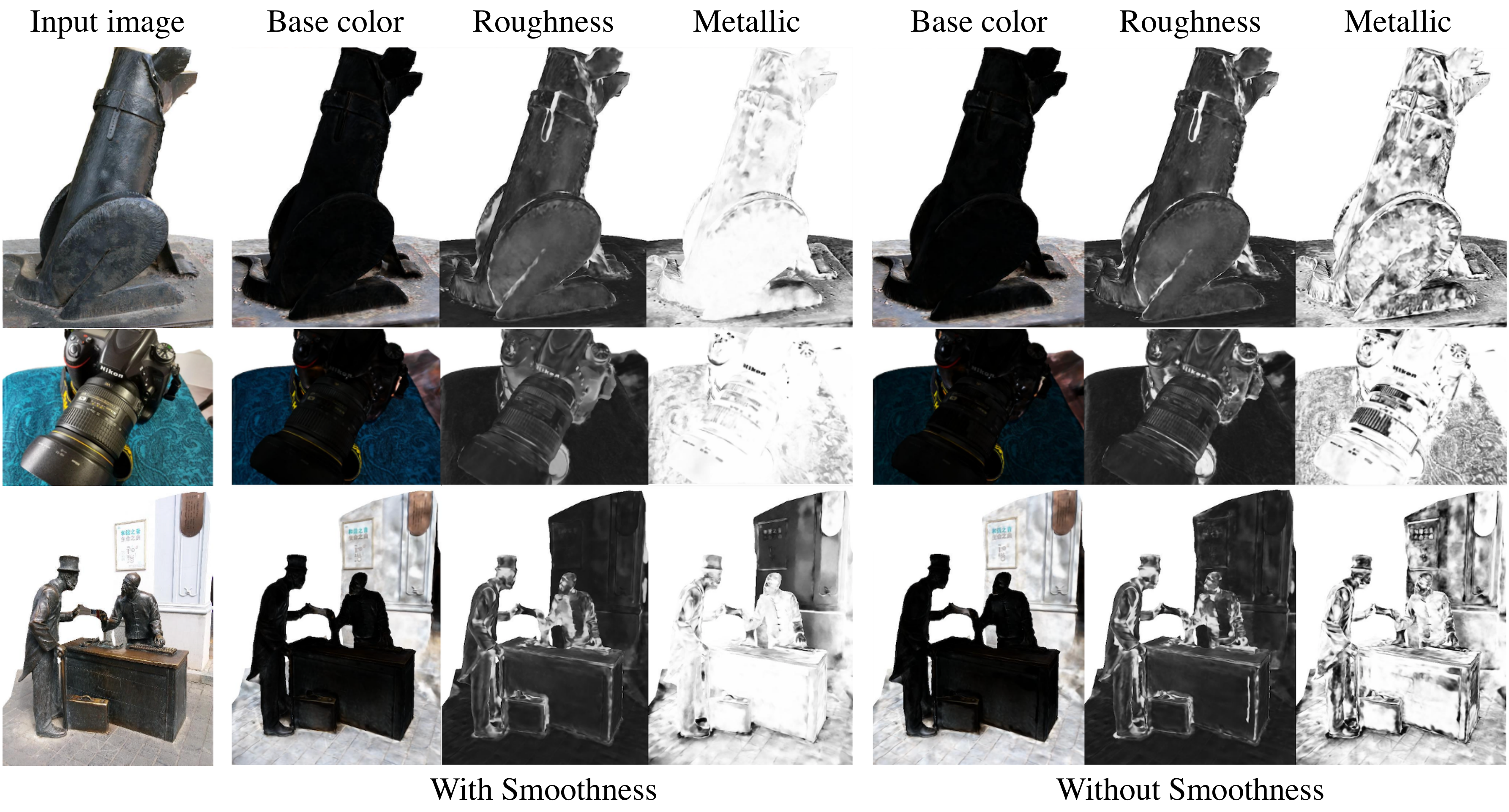}
    \captionof{figure}{Qualitative comparisons on with and without the proposed bilateral smoothness regularization.}
    \label{fig:ablation}
\end{figure}

\subsection{Ablation Study}
In this section, we analyze several design choices of the proposed framework. The ablation studies are conducted on the synthetic dataset, and we report the average scores over all scenes to compare different settings. 

\fakepara{Ray Sample Number} We first study the influence of the ray sample number for material estimation quality. The ray sample number is decreased from $S_{L} = 128$ to $S_{L} = 64$ and $S_{L} = 32$. As shown in Table~\ref{tab:ablation}, higher sampling number will lead to better reconstruction results. In our default setting, we choose $S_{L} = 128$ to better balance the quality and the VRAM/runtime consumption.

\fakepara{Random Sample} Next, we compare the fixed Fibonacci sample described in Sec.~\ref{sec:sample} with the random uniform sample commonly used in computer graphics. It is shown in Table~\ref{tab:ablation} that the random sample would lead to worse results, showing that it is crucial to precisely discretize the rendering equation in the differentiable rendering.


\fakepara{Regularizations} Lastly, we study the effectiveness of the two regularizations proposed in Sec.~\ref{sec:regularization}. We find that the bilateral smoothness is essential for material estimation of real-world scenes, where the roughness and metallic will be significantly improved if the smoothness is applied (Fig.~\ref{fig:ablation}).

\vspace{-6mm}
\begin{table}[h]
  \begin{minipage}[h]{.55\linewidth}
\normalsize \hspace{4mm} On the other hand, we also find that the two heuristics have limited influence to quantitative results of the synthetic dataset. We believe this is because the vanilla NeILF already produces high-quality estimations for synthetic scenes. In our default setting, we keep the two regularizations for all scenes but we encourage users to selectively apply the two terms depending on different characteristics of input scenes.
  \end{minipage}\hfill
  \begin{minipage}[h]{.42\linewidth}
    \centering
    \resizebox{1\textwidth}{!}{
    \begin{tabular}{c|cccc}
    \specialrule{.2em}{.1em}{.1em}
                  & Base.     & Meta.       & Roug.      & Rend.      \\ \hline
    S = 128       & \textbf{16.30} & \textbf{17.22} & 16.49          & \textbf{32.09} \\
    S = 64        & 15.20          & 16.47          & 18.88         & 31.40          \\
    S = 32        & 13.40          & 16.27          & \textbf{19.49}          & 30.57          \\ \hline
    Rand. Samp.     & 12.45	        & 15.11	        & 18.10	         & 29.73            \\
    \specialrule{.2em}{.1em}{.1em}
    \end{tabular}
    }
    \caption{Ablation studies. Average scores among all synthetic scenes are reported.}
    \label{tab:ablation}
  \end{minipage}
\end{table}

\section{Discussions}

\subsection{Comparison with NeRF Optimization}

In this section, we compare the proposed NeILF framework with the neural radiance field~\cite{martin2021nerf} optimization. We show that the two frameworks share similarities in many aspects, and thus provide readers an intuitive explanation why the proposed NeILF can successfully disentangle the complex material and lighting in the joint optimization.

\fakepara{Lighting Representations}
NeRF~\cite{martin2021nerf} represents the scene appearance as the neural radiance field. While the radiance field is physically different with the incident light field, its complexity is completely the same as ours: both NeILF and NeRF take a 3D position $\textbf{x}$ along with a direction $\boldsymbol{\omega}$ as inputs, and returns a RGB value as output. 

\fakepara{Spatially-varying Properties}
NeILF aims to recover surface materials as BRDF properties, while NeRF jointly optimizes the scene geometry as a density field. Both our BRDF and NeRF's density MLPs take only a 3D position $\textbf{x}$ as input, and return different spatial properties as outputs. Implementation-wise, the only difference is that our BRDF is consist of a 5D parameter vector, while the density value is a 1D scalar. Nevertheless, the two spatially-varying properties are very similar and their complexities are comparable. 

\fakepara{Rendering Formulations}
Our method applies the physically-based rendering to compute the reflected light from a surface point, while NeRF adopts the volume rendering to get the accumulated color along a viewing ray. On the one hand, NeILF requires incident light integration over the hemisphere; on the other hand, NeRF requires alpha composition along the viewing ray. Implementation-wise, to render a pixel, NeILF needs to sample the BRDF MLP once and the incident light MLP for multiple times, while NeRF does the same operations on the density and radiance MLPs.

\fakepara{Reconstruction Ambiguities}
The geometry-appearance ambiguity is addressed in NeRF++~\cite{zhang2020nerf++}. Similarly, we analyze the material-lighting ambiguity in our NeILF optimization. It has also been reported that with proper regularizations on density~\cite{oechsle2021unisurf} or converted SDF~\cite{wang2021neus,yariv2021volume}, the NeRF framework is able to produce high-quality geometry reconstructions. In contrast, we also show that the proposed bilateral smoothness can significantly improve the roughness and metallic quality for real-world scenes (Fig.~\ref{fig:ablation}). 

\subsection{Limitations and Future Works}

While the proposed method has already shown promising results for material estimation, the current pipeline still contains several limitations that could be further addressed in future works.

\fakepara{Running Speed}
Similar to NeRF, our method requires multiple samples of the light MLP to render one pixel, which makes the training process time consuming. Our current implementation takes around 1.5 hours to estimate the BRDF of a given scene (details in Sec.~\ref{sec:training}). We hope that in the future explicit Octree~\cite{yu2021plenoctrees}, spherical harmonics~\cite{Wizadwongsa2021NeX,yu2021plenoxels} or neural hashing~\cite{mueller2022instant} could be applied to speed up the NeILF optimization.

\fakepara{Geometry Optimization}
Our method assumes that the geometry of the scene should be given in advance. Although we have shown that multi-view reconstructed meshes are qualified enough for real-world DTU and BlendedMVS scenes, it would be better if we could jointly optimize the geometry during training. Possible directions include displacement/normal map estimation and recent differentiable surface refinements~\cite{yariv2020multiview,zhang2020physg,zhang2021learning}.

\section{Conclusions}

We have presented a differentiable rendering framework for material and lighting estimation. Compared with the environment map approximation, the proposed neural incident light field is capable of modelling the lighting condition of any static scenes, making it possible to estimate qualified material properties even for scenes with complex lightings and geometries.
The proposed method strictly follows the physically-based rendering equation, and jointly optimizes material and lighting through the differentiable rendering process.
We have intensively evaluated our method on our in-house synthetic dataset, the DTU MVS dataset, and the real-world BlendedMVS scenes. Our method is able to outperform previous environment map based methods by a significant margin in terms of the novel view rendering quality, setting a new state-of-the-art for image-based material and lighting estimation.

\clearpage
%
%
\bibliographystyle{splncs04}
\bibliography{egbib}

\newpage

\begin{center}
    \vspace{2mm}
    \textbf{\LARGE Supplementary Material for NeILF}
\end{center}
\setcounter{section}{0}
\setcounter{equation}{0}
\setcounter{figure}{0}
\setcounter{table}{0}
\setcounter{page}{1}

\vspace{2mm}
    
\noindent In this supplementary material, we show implementation details and additional results of the proposed method. For relighting results, please refer to our \textbf{supplementary video}.

\section{Implementation Details}

\subsection{Network Architecture}
\begin{figure}[h]
    \centering
    \includegraphics[width=0.9\linewidth]{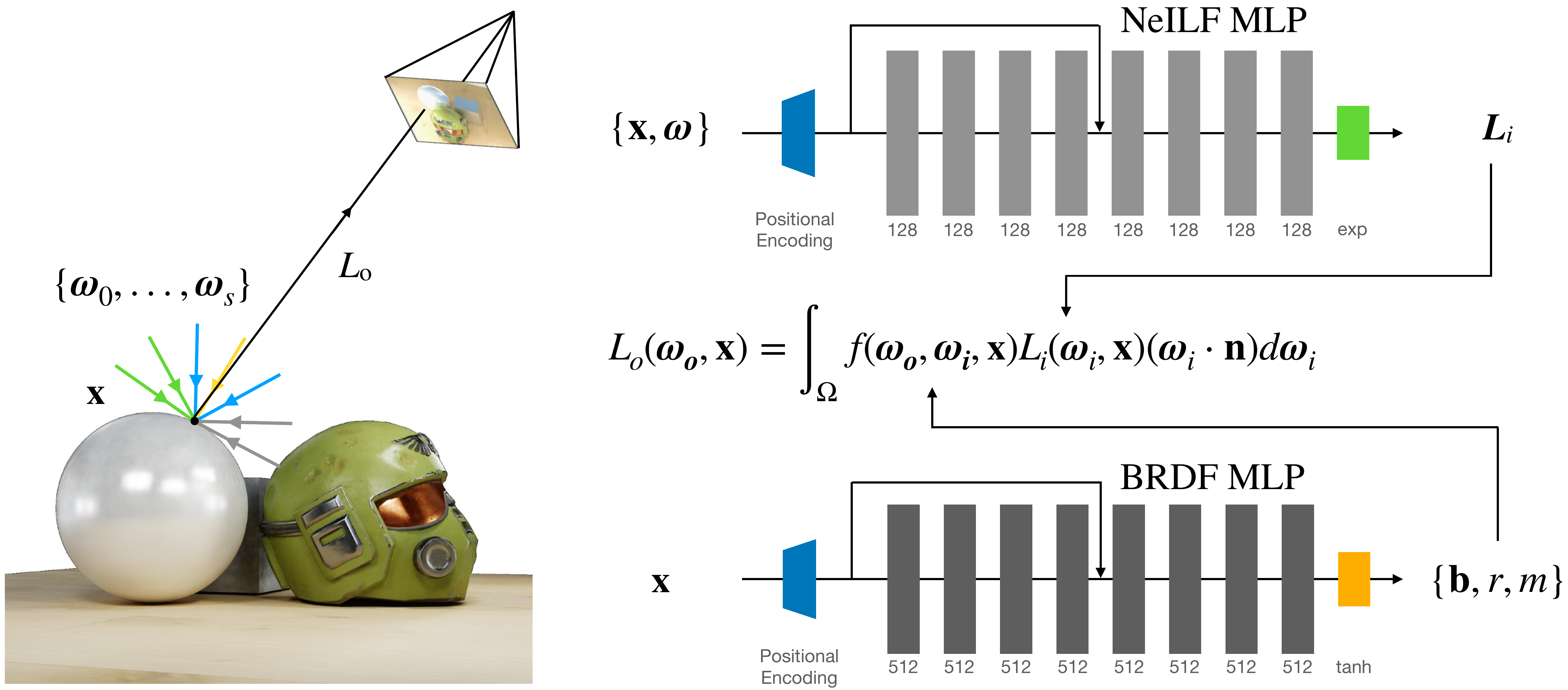}
    \caption{Network architecture of the proposed NeILF.}
    \label{fig:network}
\end{figure}

The detailed architecture of the proposed method is shown in Fig.~\ref{fig:network}. The overall design of our network is simple yet effective. We believe the architecture can be easily re-implemented or extended by other researchers.

\vspace{-5mm}
\subsection{Detailed BRDF Functions}

In this section, we describe detailed implementations of our normal distribution term D, Fresnel term F, and geometry term G. The normal distribution function D is approximated by the Spherical Gaussian function:
\begin{equation*}
	\begin{aligned}
		D(\textbf{h}; r) = S(\textbf{h}, \frac{1}{\pi r^2}, \textbf{n}, \frac{2}{r^2}) = \frac{1}{\pi r^2}e^{\frac{2}{r^2} (\textbf{h} \cdot \textbf{n} - 1)}.
	\end{aligned}
\end{equation*}
The Fresnel term is given as:
\begin{equation*}
    \begin{aligned}
        F(\boldsymbol\omega_o, \mathbf{h}; \mathbf{b}, m) &= F_0 + (1-F_0)(1-(\boldsymbol\omega_o \cdot \mathbf{h})^5), \\
        \text{where } F_0 &= 0.04(1-m) + \mathbf{b}m.
    \end{aligned}
\end{equation*}
Finally, the geometry term is approximated by the GGX function~\cite{walter2006ggx}:
\begin{equation*}
\begin{aligned}
G(\boldsymbol{\omega}_i, \boldsymbol\omega_o, \mathbf{n}; r) &= G_{GGX}(\boldsymbol\omega_i \cdot \mathbf{n})G_{GGX}(\boldsymbol\omega_o \cdot \mathbf{n}), \\
\text{where } G_{GGX}(z) &= \frac{2z}{z+\sqrt{r^2+(1-r^2)z^2}}.
\end{aligned}
\end{equation*}

\section{Synthetic Dataset}

\subsection{Dataset Setup} 

\begin{figure}[h]
  \begin{minipage}[h]{.50\linewidth}
    \centering
    \includegraphics[width=0.88\linewidth]{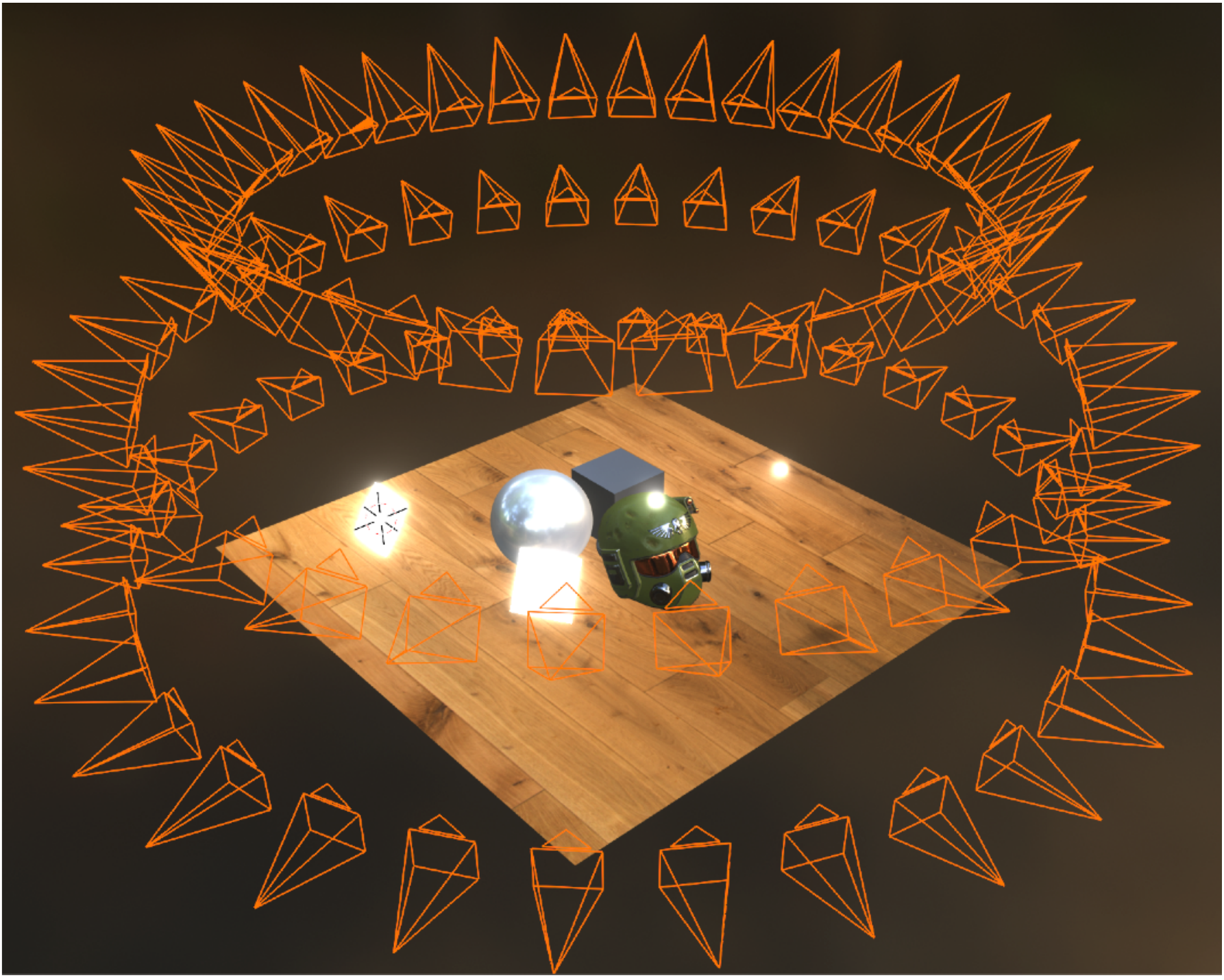}
    \caption{The camera and the near-range light setting of our synthetic dataset. We add two point lights and two area lights in the mixed lighting to lit the scene.}
    \label{fig:setup}
  \end{minipage}\hfill
  \begin{minipage}[h]{.46\linewidth}
    \centering
    \includegraphics[width=1\linewidth]{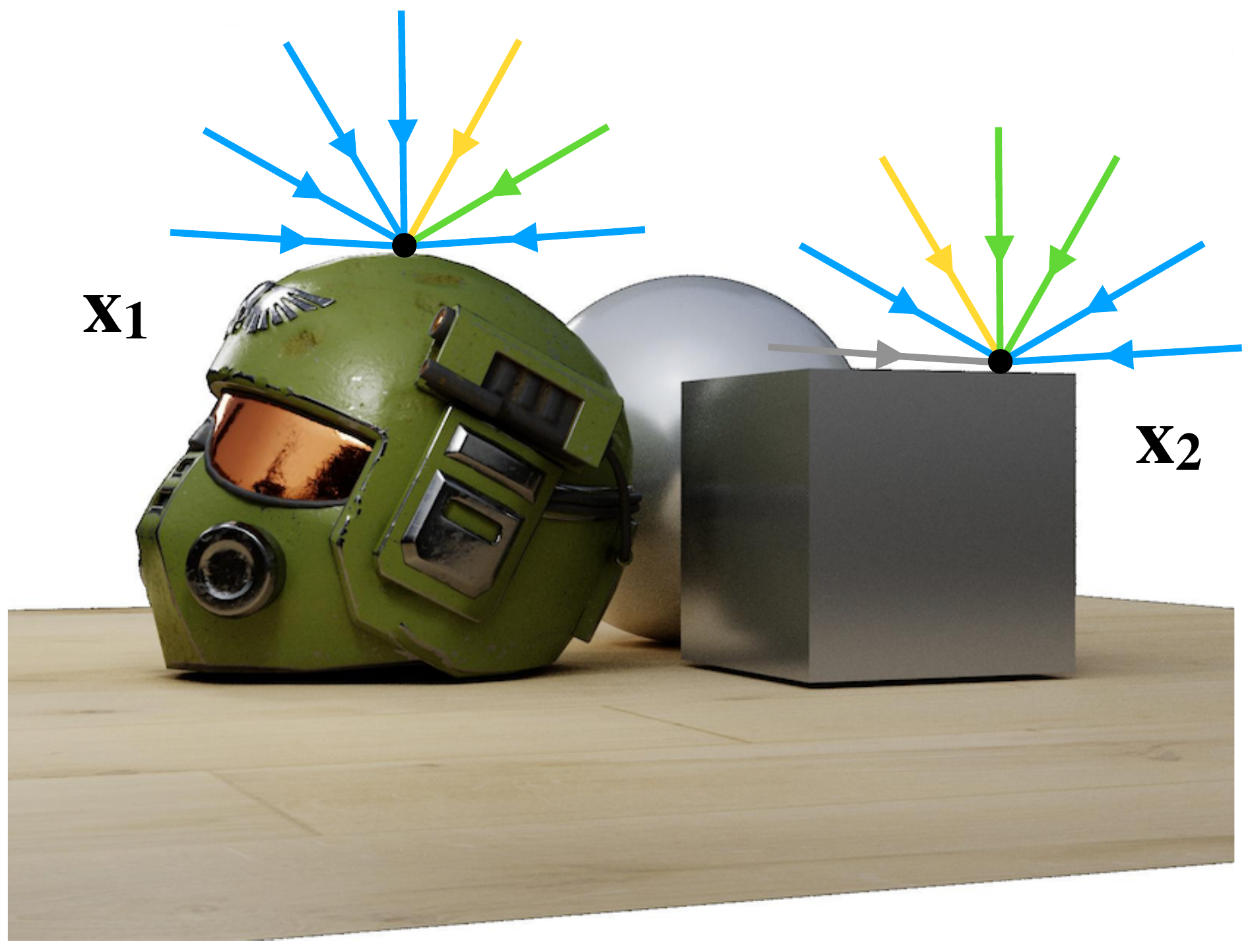}
    \caption{Two selected points for incident light visualization. Estimated incident lights for $\mathbf{x}_1$ and $\mathbf{x}_2$ are visualized in Fig.~\ref{fig:lighting}}
    \label{fig:points}
  \end{minipage}
\end{figure}

In this section, we illustrate the camera and lighting setup of our in-house synthetic dataset. The camera trajectory is shown in Fig.~\ref{fig:setup}, which contains 96 camera positions with the outside-look-in trajectory. The mixed lighting set up is also shown in Fig.~\ref{fig:setup}, which consists of two near-range point lights, two near-range area lights and one background environment map from $\{$\textit{Env-city}, \textit{Env-studio}, \textit{Env-castel}$\}$ (see main paper Sec. 5.2 for details).  

\vspace{-3mm}
\subsection{Lighting Estimation} 

\begin{figure}[H]
    \centering
    \includegraphics[width=1.0\linewidth]{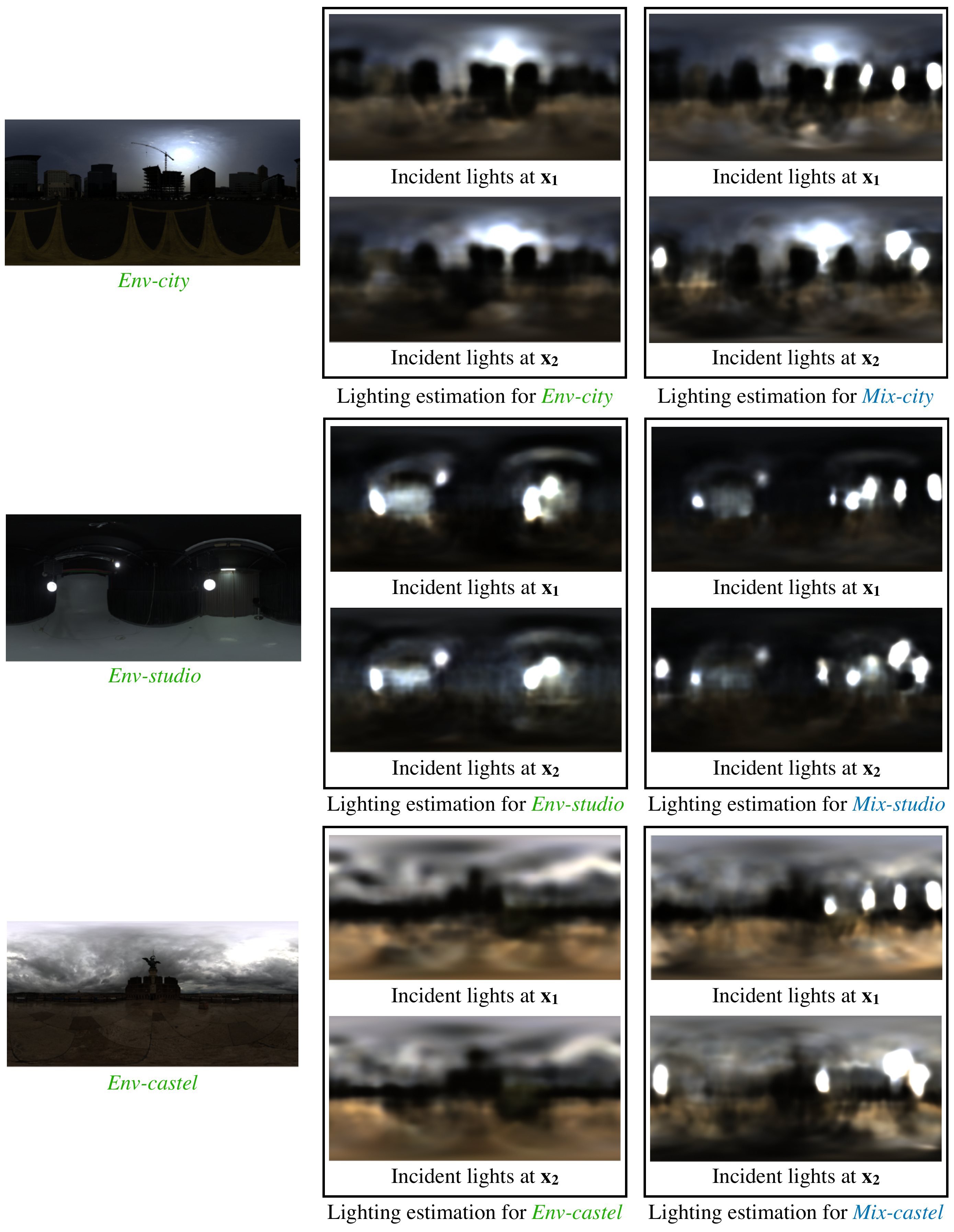}
    \caption{Visualizations of our lighting estimations. For environment map based lightings (\textit{Env-city}, \textit{Env-studio} and \textit{Env-studio}), our incident light estimations correctly recovery GT environment maps at both points (see Fig.~\ref{fig:points}). For mixed lightings with point and area lights (\textit{Mix-city}, \textit{Mix-studio} and \textit{Mix-studio}), our results well explain all light sources. Note that the three mixed lightings share the same near-range light settings as shown in Fig.~\ref{fig:setup}, and we are able to generate consistent point and area light estimations (high lights in images) across different background environments.}
    \label{fig:lighting}
    \vspace{-5mm}
\end{figure}

Due to the space limit, we did not analyze the lighting estimation result in the main paper. In this section, we demonstrate the powerful capability of the proposed NeILF for lighting modelling. As shown in Fig.~\ref{fig:points}, we select two surface points on top of the helmet and the cube to show the estimated incident lights at these two points. The lighting estimation results under six lighting conditions are visualized in Fig.~\ref{fig:lighting}. Our lighting estimations enjoy the following properties:
\begin{itemize}
    \item For environment map based lightings (\textit{Env-city}, \textit{Env-studio} and \textit{Env-studio}), our incident light estimations successfully recovery corresponding environment maps at both points.
    \item For mixed lightings (\textit{Mix-city}, \textit{Mix-studio} and \textit{Mix-studio}), our results well explain all light sources, including background environment maps, two near-range point lights, and two near-range area lights.
    \item For mixed lightings, we generate consistent point and area light estimations (high lights in images) across different background environments.
\end{itemize}

\section{DTU and BlendedMVS Datasets}

\begin{figure}[h]
    \centering
    \includegraphics[width=1\linewidth]{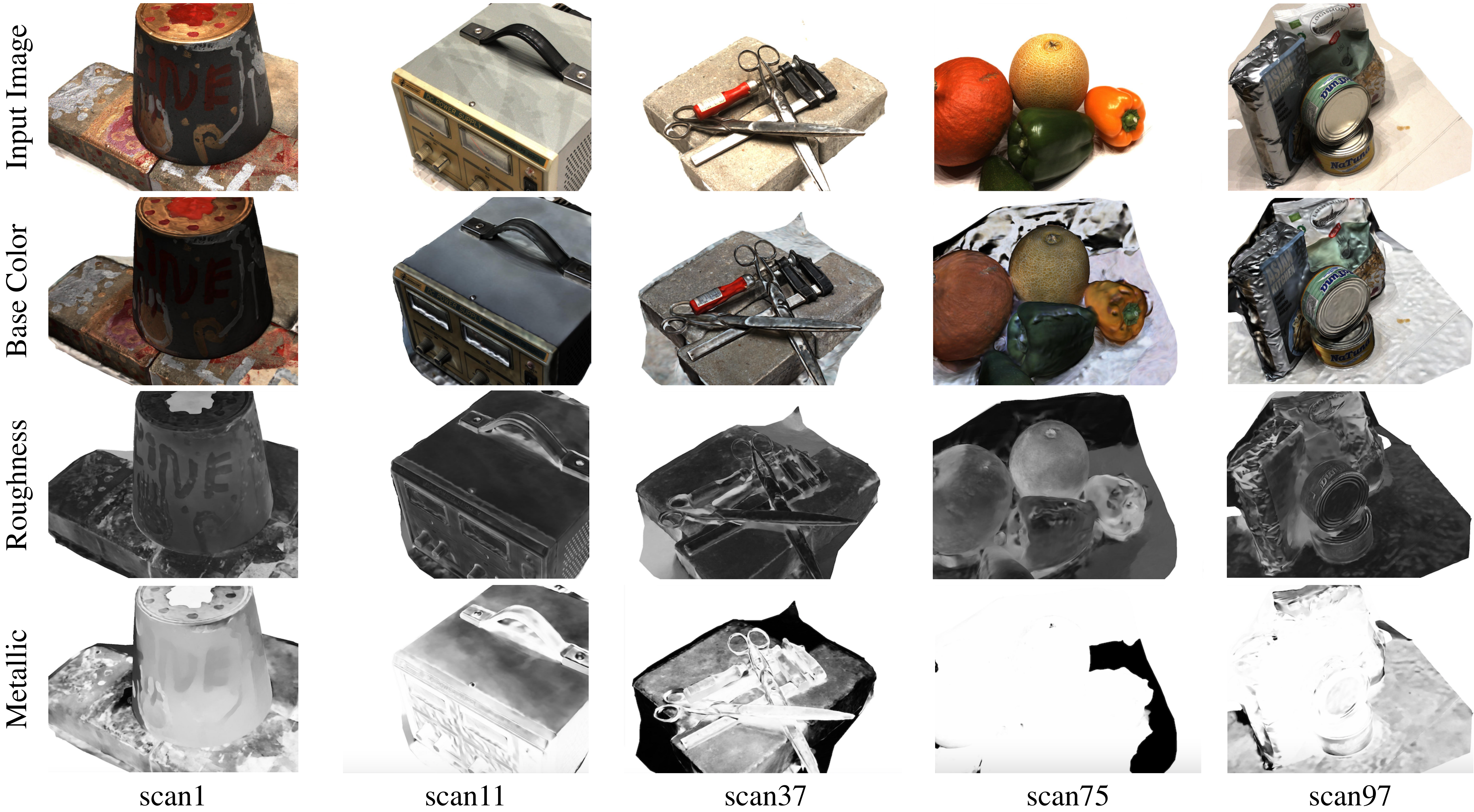}
    \caption{Results of DTU~\cite{jensen2014large} scenes.}
    \label{fig:dtu_all}
    \vspace{-10mm}
\end{figure}

\begin{figure}[h]
    \centering
    \includegraphics[width=1\linewidth]{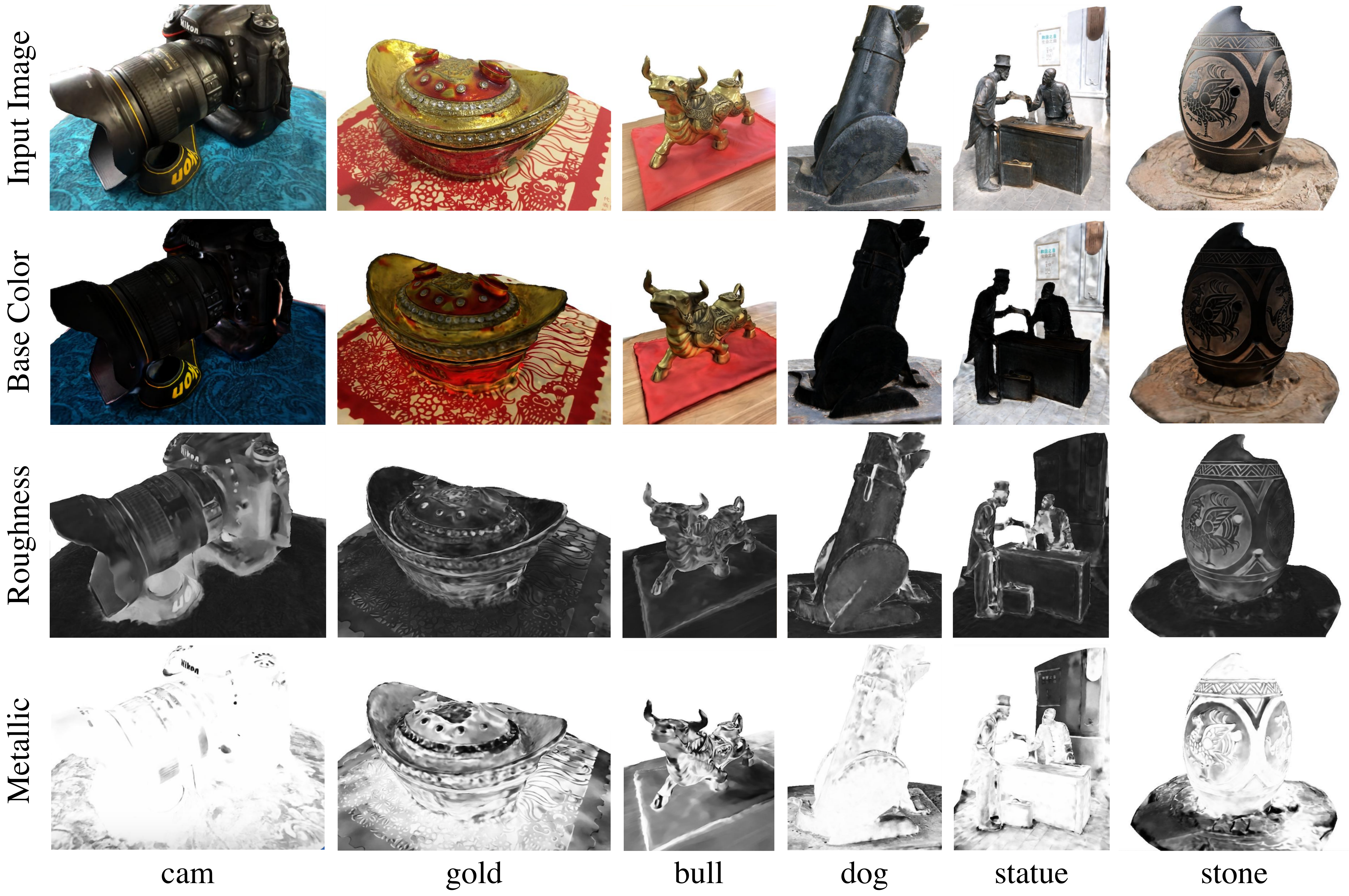}
    \caption{Results of BlendedMVS~\cite{yao2020blendedmvs} scenes.}
    \label{fig:blended_all}
    \vspace{-25mm}
\end{figure}

We show BRDF estimation of all DTU~\cite{jensen2014large} scenes in Fig.~\ref{fig:dtu_all} and all BlendedMVS~\cite{yao2020blendedmvs} scenes in Fig.~\ref{fig:blended_all}. 

\vspace{-2mm}

\newpage

\section{Relighting}

\begin{figure}[h]
    \centering
    \includegraphics[width=1\linewidth]{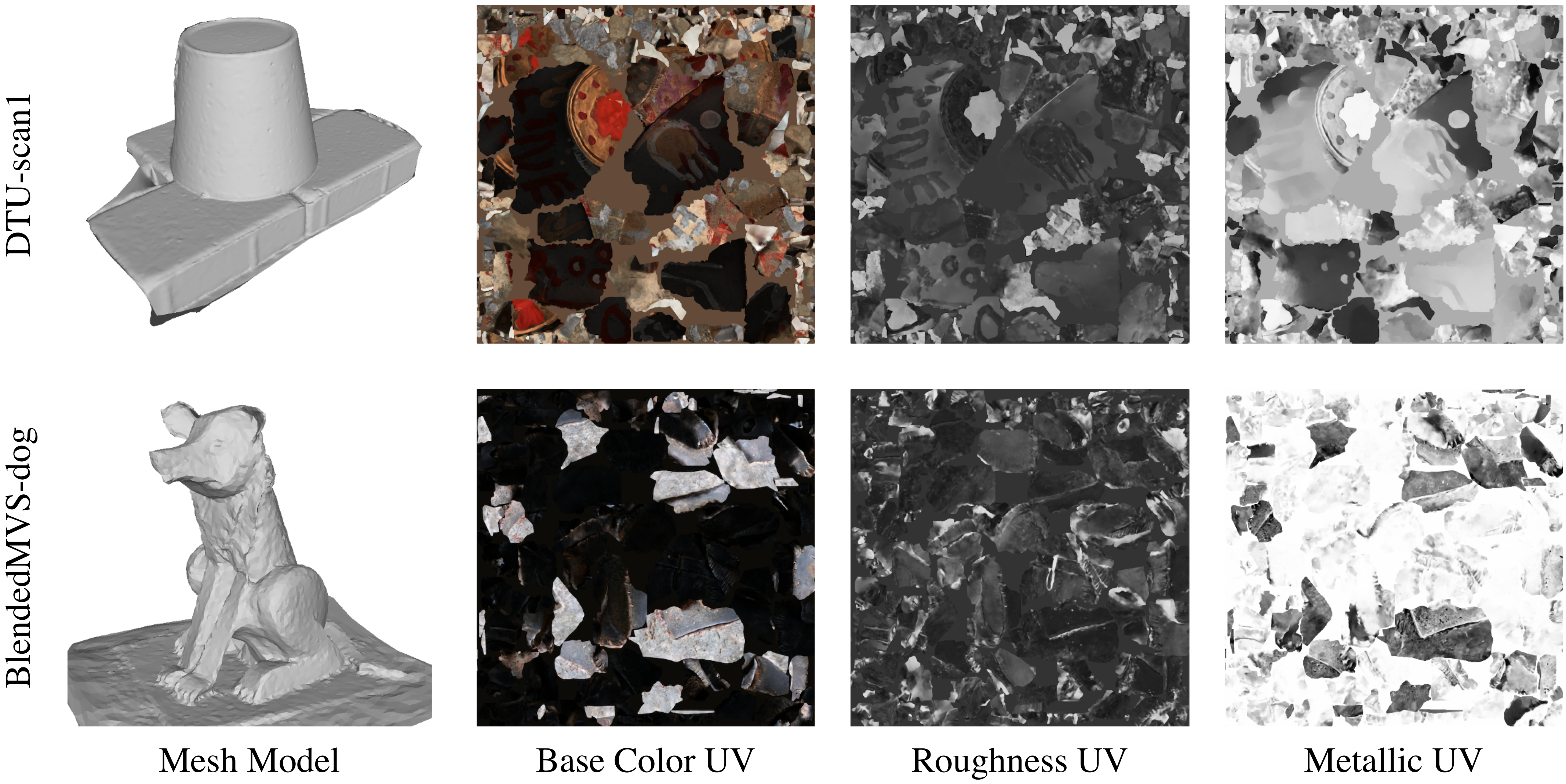}
    \caption{Exported BRDF UV maps.}
    \label{fig:uv}
    \vspace{-30mm}
\end{figure}

In this section, we show the relighting result of our BRDF estimation. Given a mesh geometry input, we apply the the Iso-charts\cite{zhou2004iso} algorithm to paramiterize the mesh surface as a 2D UV map. Then, for each pixel in the UV map, we pass the corresponding surface point to the BRDF MLP to obtain its BRDF values. The resulting BRDF texture maps (Fig.~\ref{fig:uv}) can be directly used in rendering pipelines for relighting. Please refer to our \textbf{supplementary video} for convincing results.

\end{document}